\documentclass[letterpaper, 10 pt, conference]{ieeeconf}  

\IEEEoverridecommandlockouts                              

\usepackage{graphics} 
\usepackage{epsfig} 
\usepackage{mathptmx} 
\usepackage{times} 
\usepackage{amsmath} 
\usepackage{amssymb}  
\usepackage{algorithm,algorithmic}
\usepackage[hidelinks]{hyperref}

\title{\LARGE \bf
Adaptive Memory and Reflection Multi-Agent System for Medical Question Answering
}

\author{Pradeep Murugesan$^{1}$, Luoxiao Yang$^{2}$, Xueli Chen$^{3}$, Xinqi Fan$^{1*}$
\thanks{$^{1}$Pradeep Murugesan and Xinqi Fan are with the School of Computing and Mathematics, Manchester Metropolitan University, UK. 
{\tt\small pradeep71195@gmail.com, x.fan@mmu.ac.uk}
}%
\thanks{$^{2}$Luoxiao Yang is with the Electrical and Computer Engineering, Technion – Israel Institute of Technology, Israel. 
}%
\thanks{$^{3}$Xueli Chen is with the School of Science and Technology, Hong Kong Metropolitan University, Hong Kong SAR. 
}%
\thanks{This work was partially supported by a grant from the Research Grants Council of Hong Kong SAR (No. UGC/FDS16/E16/25). }
\thanks{$^{*}$Corresponding author}
}

\begin{document}

\maketitle
\thispagestyle{empty}
\pagestyle{empty}


\begin{abstract}
Accurate and responsible medical question answering (QA) is important in healthcare, where complex cases require factual knowledge and nuanced reasoning. Existing medical QA systems, typically based on single-agent architectures and static retrieval, often lack adaptability, persistent memory, and structured decision-making. This work introduces an adaptive memory and reflection (AMR) agentic system, a multi-agent framework in which specialized agents use dedicated memory and reflection-based feedback to retrieve relevant prior cases and improve subsequent reasoning. Complexity assessment routes questions through solo, collaborative, or escalated workflows, while consensus and ethical overseer modules support reasoning consolidation and output review. 
Evaluation on MedQA and MedMCQA demonstrates strong performance compared with several baselines. Ablation studies show that combining agent-specific memory, reflection, and external retrieval yields the strongest performance. These findings highlight the potential of structured memory and feedback for developing more trustworthy medical agents.
The source code is publicly available at \href{https://github.com/mm-air/AMR-Agent}{https://github.com/mm-air/AMR-Agent}.
\end{abstract}

\section{INTRODUCTION}
Medical question answering (QA) aims to answer natural-language medical questions by grounding responses in biomedical knowledge, clinical evidence, and domain-specific reasoning~\cite{jin2022biomedical}. It supports concise evidence access at the point of care~\cite{goodwin2016medical}, biomedical literature understanding, medical education and examination-style assessment, and evaluation of expert-level reasoning in AI systems~\cite{goodwin2016medical}. The field has developed from clinical evidence retrieval systems to modern natural language processing (NLP) and large language model (LLM) settings~\cite{singhal2025toward}.

Current medical QA studies mainly proceed along three directions. The first direction explores stronger domain-specific LLMs for better medical analysis~\cite{thirunavukarasu2023large}. The second direction improves answering with retrieval-augmented generation (RAG) so that responses can be grounded in external medical evidence~\cite{sohn2025rationale}. The third direction uses agents to imitate clinical decision discussions~\cite{kim2024mdagents}. Although there are many advances in medical QA systems, they often lack persistent memory and reflection capabilities, preventing them from learning from past errors or improving reasoning over time \cite{liu2026medchain, lee2023towards, Choudhury2024}. Furthermore, the absence of explicit ethical control mechanisms contributes to unreliable outputs in clinical contexts \cite{Clusmann2023, borkowski2025multiagent}. While RAG and multi-agent approaches partially address these issues, they remain insufficient without integrated adaptability and feedback.

To address these limitations, we propose an adaptive memory and reflection (AMR) agentic system, a multi-agent medical QA framework integrating role-specific memory and a structured reflection-feedback loop. Built on graph-based orchestration \cite{LangGraph2025}, AMR supports dynamic routing, consensus-driven reasoning \cite{chen2024reconcile}, and post-generation safety screening \cite{Pham2025}. Inspired by clinical team processes \cite{Epstein2014}, this design supports more transparent and controllable medical QA workflows. The main contributions are:
\begin{itemize}
    \item \textbf{A medical QA framework based on multi-agent collaboration and clinically inspired design.} We propose an AMR agentic system, a graph-based framework that separates complexity-aware routing, collaborative reasoning, synthesis, and an ethical overseer.
    \item \textbf{Role-specific memory and reflection for adaptive reasoning.} We introduce agent-specific memories and post-hoc reflection updates so that prior cases are reused in a role-aware manner rather than through a single undifferentiated memory store.
    \item \textbf{An empirical study of component interactions in medical QA.} On MedQA~\cite{jin2021disease} and MedMCQA~\cite{pal2022medmcqa} benchmarks, we evaluate the effects of adaptive memory and reflection within the same framework and show that the combined system design yields the strongest gains.
\end{itemize}

\begin{figure*}[!t]
    \vspace{0.2cm}
    \centering
    \includegraphics[width=0.95\linewidth]{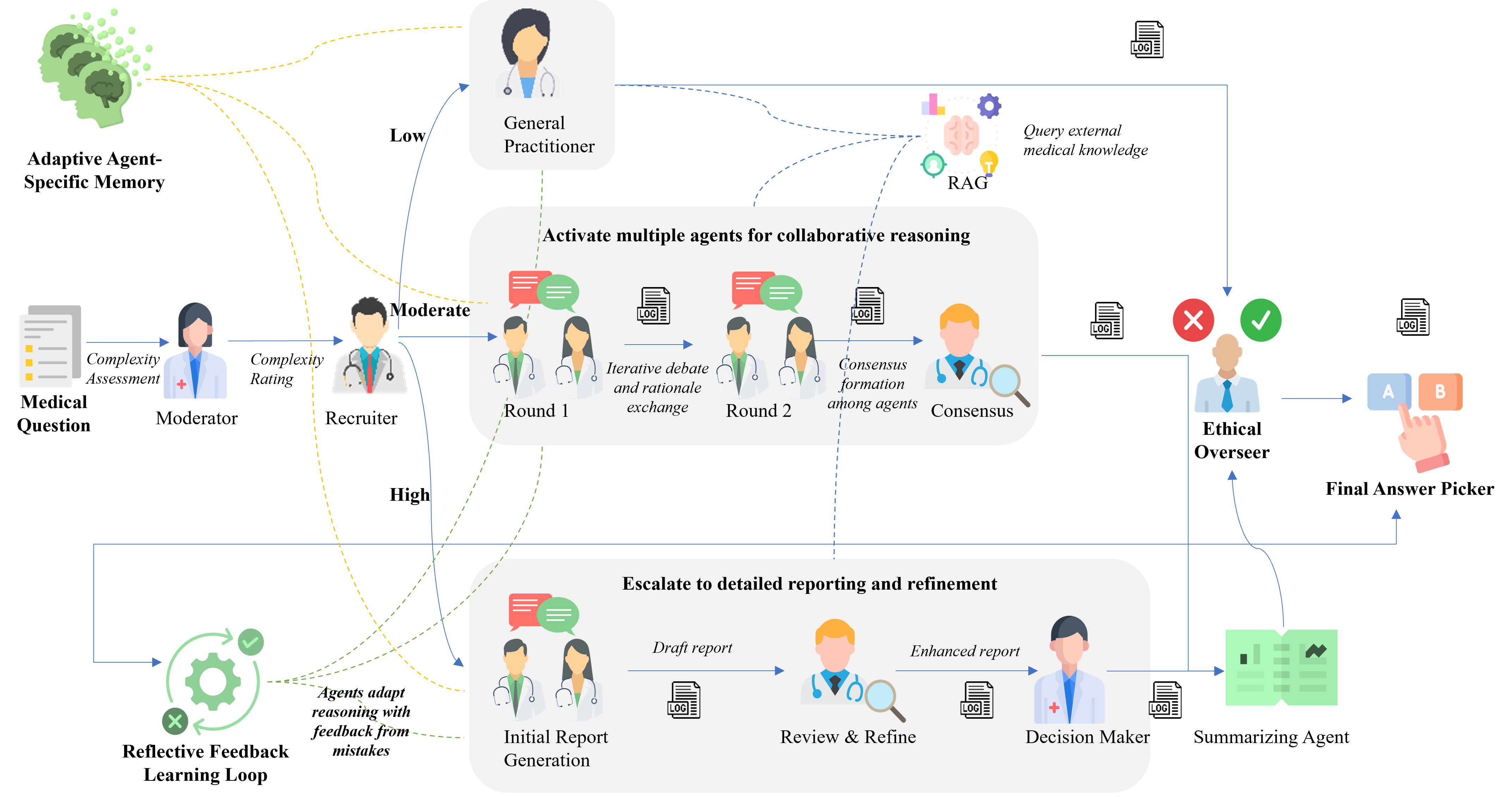}
    \caption{Pipeline of the adaptive memory and reflection (AMR) agentic system, including complexity assessment, dynamic routing, collaborative reasoning, memory, feedback, and post-generation safety screening.}
    \label{fig:architecture}
\end{figure*}

\section{RELATED WORK}
Large language models (LLMs) have significantly advanced medical NLP, outperforming traditional approaches in tasks such as clinical summarization, concept extraction, and question answering \cite{Kumar2024, Yang2020}. Domain-specific models trained on biomedical corpora further improve contextual understanding, while integration with structured ontologies enhances interpretability and reduces hallucination \cite{chang2024use, liu2026medchain}. 
Despite strong performance in clinical decision support and QA tasks, LLMs remain limited by hallucination, reasoning instability, and lack of transparency \cite{Clusmann2023}. Additionally, regulatory and privacy constraints and challenges in retrieval alignment limit real-world deployment \cite{McKeown2021, amugongo2025retrieval}.

Multi-agent systems (MAS) enable distributed, collaborative reasoning aligned with the interdisciplinary nature of clinical decision-making \cite{Shakshuki2015}. While early systems relied on rule-based coordination \cite{Epstein2014}, modern approaches employ LLM-based agents that communicate and collaborate to improve reasoning accuracy, robustness, and interpretability \cite{kim2024mdagents, zuo2025kg4diagnosis, borkowski2025multiagent}. However, challenges remain in managing agent disagreement, coordination latency, and adaptive collaboration strategies \cite{johnson2025exploring}.
Retrieval-augmented generation (RAG) improves factual grounding by integrating external knowledge during inference \cite{Singh2025, Gargari2025}. While effective in enhancing accuracy, RAG systems remain sensitive to retrieval misalignment, which can degrade reasoning consistency \cite{Selbie2024}. Moreover, clinical usability aspects such as explainability and evidence traceability are still underexplored.

In contrast to the above paradigms, the proposed AMR framework introduces two system-level shifts in medical QA design. Firstly, AMR treats medical QA as a pipeline design problem and makes routing explicit, so that reasoning depth is selected according to question complexity rather than fixed for every case. Secondly, AMR combines role-specific memory, post-hoc reflection, and output screening (Ethical Overseer) in the same framework, enabling prior cases to be reused in a targeted way and candidate answers to be screened for potentially unsafe content before release. The resulting system is therefore not only a multi-agent medical QA model, but also an explicit study of how system organization changes medical QA performance.

\section{METHOD}
This section presents the proposed adaptive memory and reflection (AMR) agentic system, designed to address inconsistent reasoning, lack of persistent memory, and absence of structured feedback in LLM-based medical QA. The framework combines agent-specific memory with reflection-driven feedback to support continual learning and improved reasoning over time.

\subsection{Overall Architecture}
The AMR framework (Fig.~\ref{fig:architecture}) adopts a modular multi-agent design built on LangGraph for graph-based orchestration \cite{LangGraph2025}. Nodes represent agents or processing steps, and edges define transitions. A \textit{Moderator} assesses question complexity, after which a \textit{Recruiter} routes the query to one of three pathways: (i) \textbf{low complexity}—handled by a \textit{General Practitioner}, (ii) \textbf{moderate complexity}—resolved via multi-agent collaboration and consensus, and (iii) \textbf{high complexity}—processed through iterative hierarchical agents refinement followed by report summary and finally selection by a more senior \textit{Decision Maker}, like a lead consultant.

The implementation follows a LangGraph-style execution pipeline. The Moderator first assesses question complexity and triages the incoming query. The Recruiter activates the corresponding agents according to the assigned complexity level. The reasoning outputs are consolidated within the selected pathway and reviewed by the Ethical Overseer. The Final Answer Picker then maps the screened output to the returned answer option.

\subsection{Adaptive Routing and Collaborative Reasoning}
AMR uses different reasoning depths for different questions. Direct recall or simple mechanism questions use the low-complexity path; questions combining symptoms, findings, and specialty interpretation use specialist collaboration; and cases with competing diagnostic or treatment interpretations follow a longer workflow with draft generation, critique, refinement, and final decision. The Moderator assigns one of three routing labels (low, moderate, or high), determining the downstream execution path.

This routing avoids applying the same multi-agent procedure to every case. For moderate-complexity questions, specialists generate independent opinions, refine them using the shared transcript, and a Consensus Facilitator summarizes agreements and disagreements. For high-complexity questions, intermediate reports allow reasoning to be revised before final answer selection. Routing is therefore both a computational shortcut and a control mechanism aligning reasoning depth with question difficulty.

\subsection{Agent-Specific Memory and Reflective Update}
Each AMR agent maintains a separate memory rather than a single shared store. Memory entries contain the question context, answer, post-hoc notes, role metadata, and timestamp for later retrieval. When reflection is enabled, incorrect predictions generate an additional role-specific reflection entry containing corrective feedback. This design allows different roles to retrieve different prior experience for the same question; for example, hepatology and nephrology agents may retrieve different cases because they reason from different specialty perspectives.

In addition to memory retrieval, AMR includes a reflection loop triggered only when the predicted answer differs from the ground-truth label. Corrective feedback and reasoning summaries are stored as reflective memory to improve future retrieval rather than updating model parameters. Algorithm~\ref{alg:feedbackloop} summarizes this process.

\begin{algorithm}
\caption{Agentic Reflective Feedback Learning}
\begin{algorithmic}[1]
\FOR{each \textit{Question} in Dataset}
    \STATE \textit{Answer} $\leftarrow$ AMRSystem.process(\textit{Question})
    \STATE \textit{GroundTruth} $\leftarrow$ get\_label(\textit{Question})
    \IF{\textit{Answer} $\neq$ \textit{GroundTruth}}
        \FOR{each \textit{Agent} in team}
            \STATE \textit{Agent.memory.add(FeedbackEntry)}
        \ENDFOR
    \ENDIF
\ENDFOR
\end{algorithmic}
\label{alg:feedbackloop}
\end{algorithm}

\subsection{Answer Synthesis and Output Screening}

After specialist reasoning is complete, AMR separates answer synthesis from answer release. The \textit{Summarizer} condenses the specialist rationales into one explanation, and the \textit{Final Answer Picker} maps that explanation to the returned answer option. Before release, the \textit{Ethical Overseer} reviews the candidate response for potentially unsafe content, including direct diagnostic statements, harmful advice, or treatment recommendations that exceed the intended educational scope. Responses are labelled as APPROVED or FLAGGED, with the corresponding rationale recorded for auditability. 

The separation between synthesis and screening is important in medical QA. A candidate answer may be factually grounded yet still require filtering or abstention under a deployment policy \cite{Pham2025, McKeown2021}. By isolating output screening as an explicit stage, AMR makes policy decisions inspectable rather than burying them inside the reasoning prompt.

The Ethical Overseer functions as a policy-based post-generation screening layer that reviews candidate responses for potentially unsafe medical advice, unsupported diagnostic statements, and treatment recommendations. Rather than modifying the underlying reasoning process, it evaluates the final response against predefined safety criteria and either approves the response or flags it for non-compliance. In the current implementation, this component is intended as a preliminary safety mechanism to reduce potentially unsafe outputs and should not be interpreted as a substitute for formal clinical validation or regulatory safety assessment.

\subsection{Agent Roles and Responsibilities}
The AMR framework defines modular roles that jointly cover assessment, recruitment, reasoning, synthesis, and screening. 
AMR includes several agents with distinct responsibilities; Table~\ref{table:agent_roles} summarizes their roles and functions.

\begin{table}[!t]
\vspace{0.3cm}
\centering
\footnotesize
\setlength{\tabcolsep}{3pt}
\caption{Roles and Responsibilities of AMR Agents}
\label{table:agent_roles}
\begin{tabular}{p{2.4cm}p{2.1cm}p{3.4cm}}
\hline
\textbf{Agent Node} & \textbf{Input} & \textbf{Output / Function} \\
\hline
Moderator & User question & Assesses complexity \\
Recruiter & Complexity, question & Allocates agents \\
General Practitioner & Question & Single-agent reasoning \\
Collaborative Agents & Question & Multi-agent reasoning \\
Initial Report Agent & Question & Draft report \\
Review/Refine Agent & Report & Improves report \\
Decision Maker & Reports & Selects best output \\
Summarizer & Outputs & Consolidated rationale \\
Ethical Overseer & Answer & Flags or passes output \\
Final Answer Picker & Reports & Final selection \\
\hline
\end{tabular}
\end{table}

\section{EXPERIMENT AND DISCUSSION}
\subsection{Datasets}
We use MedQA~\cite{jin2021disease} and MedMCQA~\cite{pal2022medmcqa}, two established multiple-choice medical QA benchmarks that differ in scale, difficulty, and explanation richness. MedQA focuses on USMLE-style clinical reasoning, while MedMCQA provides large-scale exam questions with broader specialty coverage. Table~\ref{tab:datasets} summarizes key characteristics of the datasets.

For MedQA, the official training split was used to construct the retrieval corpus, while the official test split was reserved exclusively for evaluation. For MedMCQA, a disjoint subset of training questions was used for retrieval, and an independent subset was used for evaluation.

\begin{table}[!h]
    \centering
    \footnotesize
    \caption{Summary of Utilized Medical QA Datasets}
    \label{tab:datasets}
    \begin{tabular}{l|l|l}
    \hline
     & MedQA & MedMCQA \\
    \hline
    Origin & USMLE (US) & AIIMS/NEET (India) \\
    Questions & $\sim$12,700 & $\sim$194,000 \\
    Format & Clinical vignette & Multiple-choice \\
    Specialties & 21+ & 21+ \\
    Reasoning & High & Mixed \\
    Explanations & Partial & Full \\
    \hline
    \end{tabular}
\end{table}

\subsection{Implementation Details}
Experiments used a Python pipeline with JSONL inputs and GPT-4o through the OpenAI API. LangGraph orchestrated the workflow, while FAISS served as the persistent vector store. Agent memories were embedded using \textit{text-embedding-3-large} and indexed for semantic retrieval. Questions were routed according to the Moderator's complexity assessment. Agents retrieved the top five role-specific prior cases, and reflection updates were recorded after incorrect predictions. Logs captured predictions, routing decisions, retrieval traces, and ethical screening outcomes.

Each agent received role-specific prompts. The Moderator assessed complexity, the Recruiter selected specialists, specialist agents generated independent reasoning, and the Final Answer Picker returned only the final answer option. Retrieval documents were built from the training corpus, chunked into overlapping passages, embedded, and indexed in FAISS.
For computational efficiency, each benchmark was evaluated in sequential batches of 50 questions. 
To prevent test-set leakage, the retrieval corpus was constructed before evaluation and remained frozen throughout testing. Evaluation samples were processed independently without writing test questions, predictions, or reflection feedback back into persistent memory.

\subsection{Evaluation Metrics} 
We report accuracy and consistency. Accuracy measures the fraction of correctly answered questions:
\begin{equation}
\text{Accuracy} = \frac{N_C}{N} \times 100\%.
\end{equation}

Consistency is assessed by the mean $\mu$ and standard deviation $\sigma$:
\begin{equation}
\mu = \frac{1}{n}\sum_{i=1}^{n} a_i, \quad
\sigma = \sqrt{\frac{1}{n}\sum_{i=1}^{n}(a_i - \mu)^2}.
\end{equation}

\subsection{Functionality Comparison with Other Methods}
AMR introduces adaptive agent-specific memory and reflection-driven learning, supporting continual improvement and context-aware reasoning. Unlike prior multi-agent frameworks, MDAgent~\cite{kim2024mdagents}, Debate~\cite{du2024improving}, MedAgent~\cite{tang2024medagents}, ReConcile~\cite{chen2024reconcile}, which rely on static or single-pass reasoning, AMR integrates memory, feedback, and post-generation screening to support robustness and adaptability (Table~\ref{table:framework_comparison}).

\begin{table}[!ht]
\centering
\footnotesize
\setlength{\tabcolsep}{3pt}
\caption{Comparison of Functions Across Multi-Agent Methods}
\label{table:framework_comparison}
\begin{tabular}{@{}l@{\;}c@{}c@{\;}c@{\;}c@{\;}c@{\;}c@{}}
\hline
\textbf{Function} & \textbf{AMR} & \textbf{MDAgent} & \textbf{Debate} & \textbf{MedAgent} & \textbf{ReConcile} \\
\hline
Multiple Roles       & \checkmark & \checkmark & \checkmark & \checkmark & \checkmark \\
Early Stopping       & \checkmark & \checkmark & \checkmark & \checkmark & -- \\
Refinement           & \checkmark & \checkmark & \checkmark & \checkmark & -- \\
Complexity Check     & \checkmark & \checkmark & -- & -- & -- \\
Multi-party Chat     & \checkmark & \checkmark & \checkmark & -- & -- \\
Conversation Pattern  & Flexible & Flexible & Static & Static & Static \\
Ethical Overseer     & \checkmark & -- & -- & -- & -- \\
RAG Integration      & \checkmark & -- & -- & \checkmark & -- \\
Agent Memory         & \checkmark & -- & -- & -- & -- \\
Feedback Learning    & \checkmark & -- & -- & -- & -- \\
\hline
\end{tabular}
\end{table}

\subsection{Ablation Study}
We evaluate five settings: \textbf{Baseline} (multi-agent reasoning without memory, reflection, or external retrieval), \textbf{Feedback} (reflection without agent memory retrieval), \textbf{Memory} (agent-specific retrieval without reflectiion), \textbf{AMR w/o RAG} (memory and reflection), and \textbf{AMR} (full AMR augmented with external retrieval).

Table~\ref{tab:accuracy_results} reports performance across configurations. The baseline achieved 80\% (MedQA) and 78\% (MedMCQA). Feedback and memory progressively improved results, AMR without RAG reached 90\% and 87.4\%, and full AMR with RAG performed best (93.2\%, 90\%), demonstrating complementary benefits from retrieval, memory, and feedback.

\begin{table}[!ht]
\centering
\footnotesize
\caption{Ablation studies.}
\label{tab:accuracy_results}
\resizebox{\columnwidth}{!}{%
\begin{tabular}{lccc cc}
\hline
\textbf{Config} & \textbf{Mem} & \textbf{FB} & \textbf{RAG} & \textbf{MedQA} & \textbf{MedMCQA} \\
\hline
Baseline      & -- & -- & -- & 80\% & 78\% \\
Feedback      & -- & \checkmark & -- & 82\% & 80.4\% \\
Memory        & \checkmark & -- & -- & 86\% & 82\% \\
AMR w/o RAG          & \checkmark & \checkmark & -- & 90\% & 87.4\% \\
AMR      & \checkmark & \checkmark & \checkmark & 93.2\% & 90.0\% \\
\hline
\end{tabular}}
\end{table}

\subsection{Quantitative Comparison with Other Methods}
Table~\ref{tab:method_comparison} compares AMR with representative single-agent, human, and prior multi-agent results. AMR is competitive with or stronger than several automated baselines on both MedQA and MedMCQA, and approaches the reported human reference level on MedMCQA. 

\begin{table}[!ht]
\centering
\footnotesize
\caption{Comparison with Other Methods}
\label{tab:method_comparison}
\begin{tabular}{lcc}
\hline
\textbf{Method} & \textbf{MedQA} & \textbf{MedMCQA} \\
\hline
GPT-4 & 86.1\% & 73.7\% \\
Human & 87.0\% & 90.0\% \\
MDAgents & 86.5\% & 73.7\% \\
MedAgents & 83.0\% & 70.0\% \\
\textbf{AMR} & \textbf{93.2\%} & \textbf{90.0\%} \\
\hline
\end{tabular}
\end{table}

\subsection{Performance Across Different Question Complexity}
\begin{figure}[!htb]
    \centering
    \includegraphics[width=0.9\linewidth]{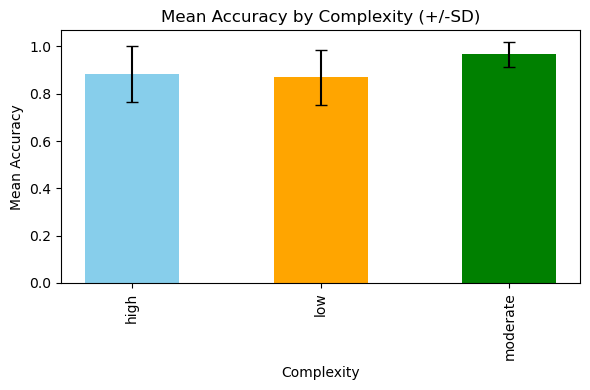}
    \caption{Accuracies across different question complexity levels.}
    \label{fig:complexity}
\end{figure}

AMR achieves high mean accuracy across all question complexity levels, with accuracy values of around 0.85 or higher (Fig.~\ref{fig:complexity}). The model achieves its highest mean accuracy and lowest standard deviation on moderate-complexity questions, which may be attributed to the multi-agent collaborative reasoning and iterative debate mechanisms. The experiments were done without enabling RAG.

\begin{figure*}[!t]
    \centering
    \vspace{0.2cm}
    \includegraphics[width=0.95\textwidth,keepaspectratio]{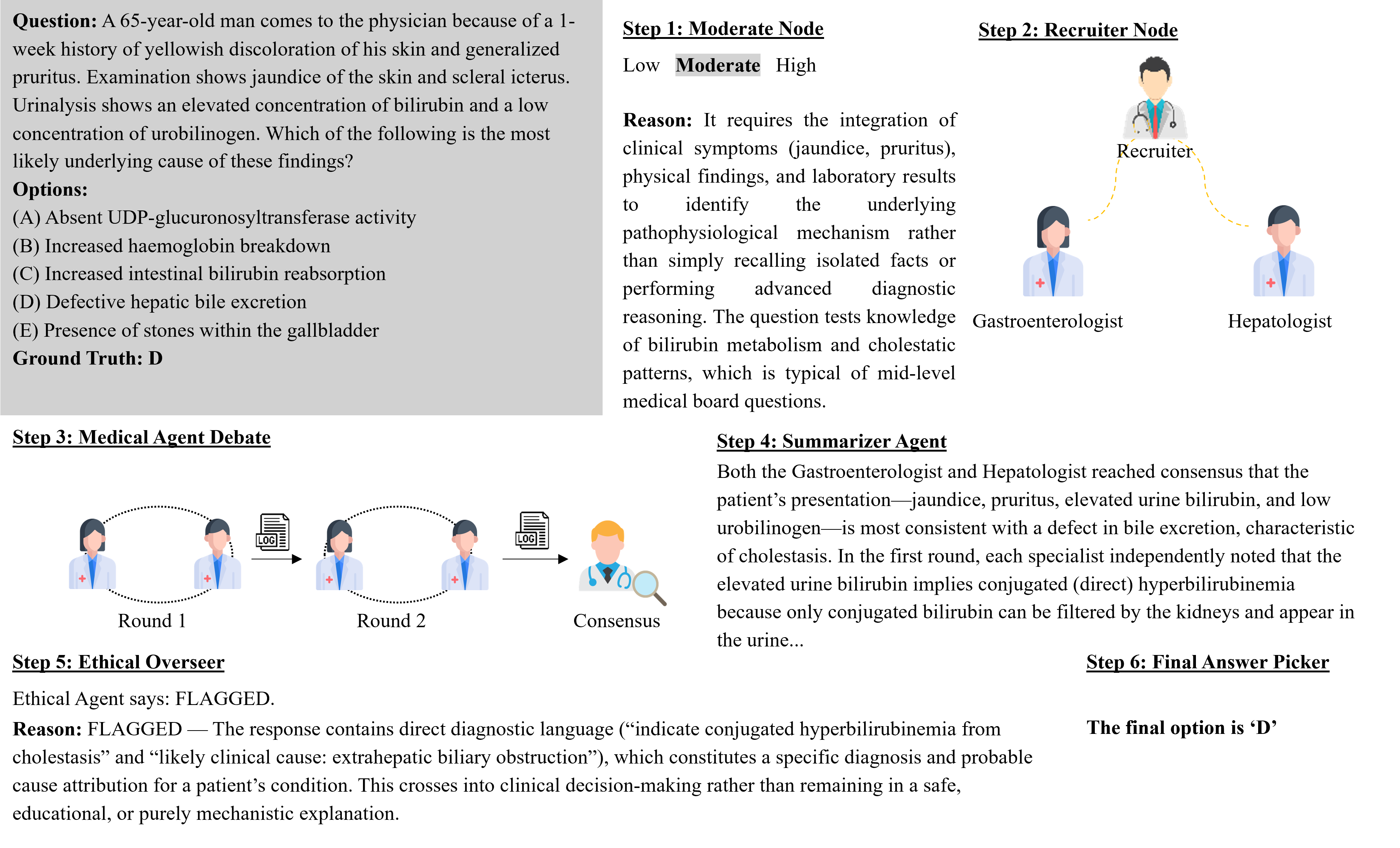}\\[-0.2cm]
    \vspace{-0.3cm}
    {\small {(a) Positive qualitative example.}}
    \vspace{0.1cm}
    \includegraphics[width=0.95\textwidth,keepaspectratio]{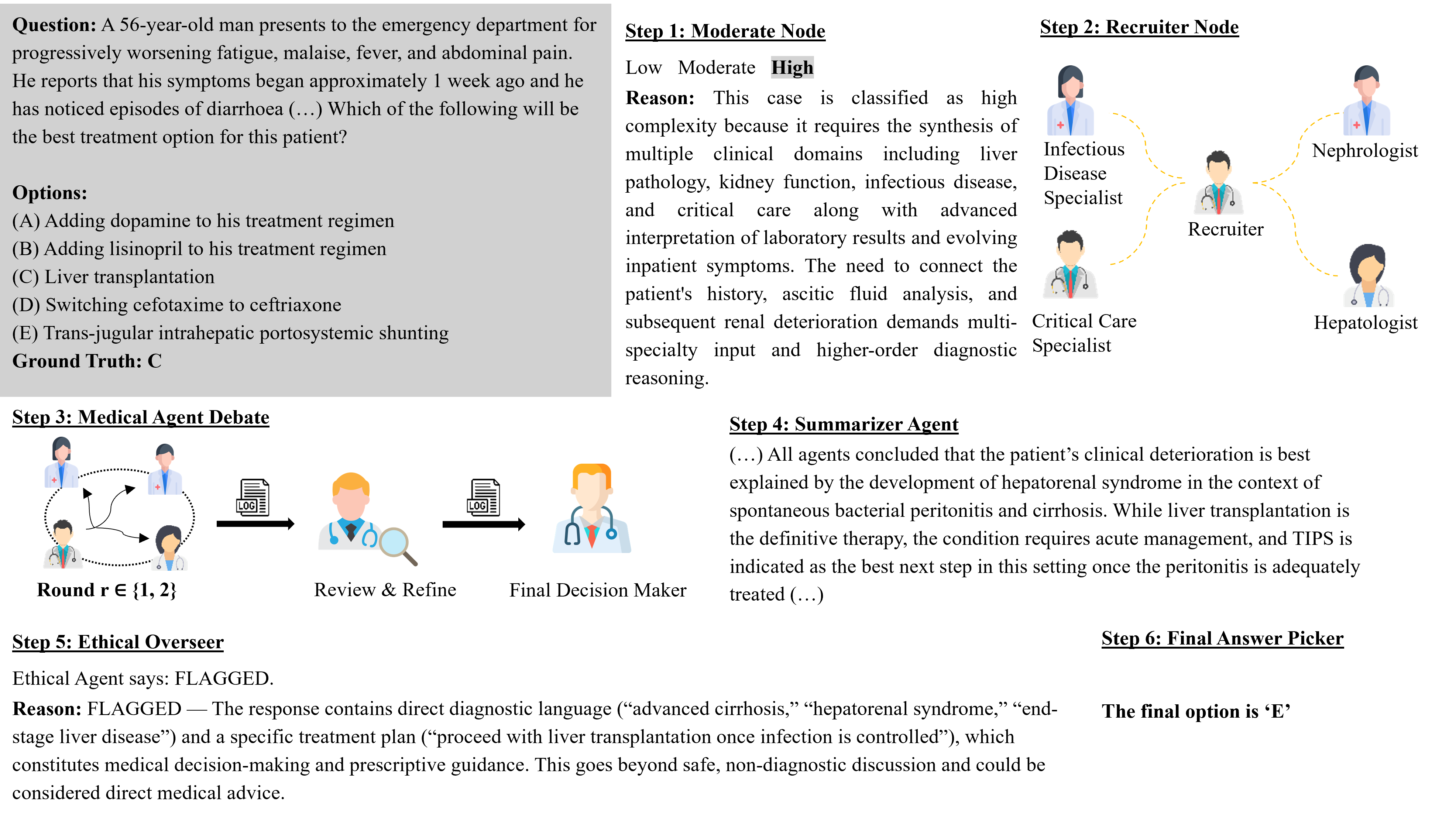}\\[-0.6cm]
    {\small {(b) Negative qualitative example.}}
    \caption{Qualitative analysis of the proposed AMR system. (a) A positive case showing correct and consistent reasoning with adaptive memory and reflection support. (b) A negative case showing an error instance used for reflection-driven feedback and future refinement.}
    \label{fig:qualitative}
\end{figure*}

\subsection{Qualitative Analysis}
Figure~\ref{fig:qualitative} presents one successful case and one failure case of the AMR system.

In the positive case (Figure~\ref{fig:qualitative}a), agents retrieve semantically similar prior cases, including surgical error disclosure scenarios, supporting consistent and context-aware reasoning across roles.

In the error case (Figure~\ref{fig:qualitative}b), the system produces an incorrect answer, exposing reasoning limitations. Such cases trigger reflection, where structured feedback (incorrect prediction, correct answer, and reasoning correction) is stored in memory. For example, the system selected option ``E'' instead of ``C'', prompting feedback on ethical disclosure prioritization. The Ethical Overseer flagged potentially unsafe recommendations and recorded the reason for review, illustrating the intended screening mechanism rather than a comprehensive quantitative evaluation.

\section{CONCLUSIONS}
This paper presents an adaptive memory and reflection (AMR) agentic framework for medical QA that integrates complexity-aware routing, agent-specific memory, collaborative reasoning, reflection, and output screening. By treating medical QA as a structured reasoning pipeline rather than a single-prompt task, AMR enables agents to collaborate, refine intermediate reasoning, and leverage both retrieved knowledge and accumulated experience. Experiments on MedQA and MedMCQA demonstrate consistent improvements over the baseline, with the combination of memory, reflection, and RAG achieving the strongest performance. 
These results demonstrate the potential of structured multi-agent systems for more accurate and adaptive medical QA.

There are several limitations. The framework is still constrained by retrieval quality, and the effectiveness of memory may decline as stored experiences grow, introducing duplicated or less relevant cases. The reflection mechanism has only been evaluated on retrospective benchmarks rather than real clinical workflows. In addition, the ethical overseer uses LLM knowledge and has not been validated using dedicated clinical rules or by clinicians. Finally, MedQA and MedMCQA do not fully capture complexity and workflow requirements of real-world clinical decision-making.

Future work will investigate more effective memory management and retrieval strategies, including memory pruning, confidence-based retention, recency-aware retrieval, reranking, and forgetting mechanisms, to improve long-term performance. We also plan to extend AMR beyond multiple-choice benchmarks to open-ended clinical reasoning and real-world decision support, while conducting systematic evaluations of safety, inference latency, token consumption, and the trade-offs between reasoning quality and computational efficiency.


\addtolength{\textheight}{0cm}   





{\small
\bibliographystyle{ieee}
\bibliography{reference}
}

\end{document}